\documentclass[letterpaper,twocolumn,fleqn]{article} 

\usepackage{ist}
\usepackage{mathtools}
\usepackage{titlesec}
\usepackage{amsmath}
\usepackage{amssymb}
\usepackage{comment}
\usepackage{subfig}
\usepackage[]{algorithm2e}
\title{Page Classification for Print Imaging Pipeline}

\author{Shaoyuan Xu\textsuperscript{ a}, Cheng Lu\textsuperscript{ a}, Mark Shaw\textsuperscript{ b}, Peter Bauer\textsuperscript{ b}, Jan Allebach\textsuperscript{ a}\\\textsuperscript{ a}School of Electrical and Computer Engineering, Purdue University, West Lafayette, IN 47906, U.S.A.\\\textsuperscript{b}HP Inc., Boise, ID 83714, U.S.A.}

\date{2016-Nov.-18} 

\begin{document}
\maketitle 

\thispagestyle{empty} 


\titlespacing*{\chapter} {0pt}{50pt}{40pt}
\titlespacing*{\section} {0pt}{3.5ex plus 1ex minus .2ex}{2.3ex plus .2ex}
\titlespacing*{\subsection} {0pt}{3.25ex plus 1ex minus .2ex}{1.5ex plus .2ex}

\begin{abstract}
Digital copiers and printers are widely used nowadays. One of the most important things people care about is copying or printing quality. In order to improve it, we previously came up with an SVM-based classification method to classify images with only text, only pictures or a mixture of both based on the fact that modern copiers and printers are equipped with processing pipelines designed specifically for different kinds of images\cite{bib1}. However, in some other applications, we need to distinguish more than three classes. In this paper, we develop a more advanced SVM-based classification method using four more new features to classify 5 types of images which are text, picture, mixed, receipt and highlight. 
\end{abstract}

\section{1.0 Introduction}
\subsection{1.1 General Introduction of the Project}

In everyday life, people have all kinds of images which they want to scan or print. That is why digital copiers or all-in-one printers are needed. And one of the most important things people care about is the printing quality of digital copiers or all-in-one printers.

In order to optimally process different kinds of input images, modern copiers and printers possess multiple processing pipelines. Each one of these pipelines are designed specifically for one type of image based on their unique features. For text images, we require the text area to have clear, sharp edges and also high contrast. But for picture images, we do not want them to have high contrast. Instead, we want them to be more blurred. If an image is copied or printed by the right pipeline, people will get the image of the best quality. However, if an image is copied or printed by the wrong pipeline, the image of significantly bad quality will be produced. For example, if a text image is processed by a picture pipeline, the text area in the output image will have blurred edges and comparably low contrast which is not satisfactory\cite{bib2}. Figure 1 illustrates an example that a text image is processed by both picture and text pipelines. It is obvious that the text image is more clear and has better edge sharpness when it is processed by a text pipeline which is more desirable. So in order to increase the copying or printing quality, we want to come up with a method to classify the images going into copiers or printers and process them in different ways. According to the previous research work done by Cheng Lu from our group, we have already had a SVM-based classification method to classify three types of images: text, picture and mixed and we have three features to train the classifiers\cite{bib1}. In this paper, we develop four new features to classify two new image classes which are highlight and receipt.

In this paper\footnote{
Research supported by HP Inc., Boise, ID 83714.
}, we develop four new features to classify two new image classes which are highlight and receipt.

For the rest of Section 1, we introduce five types of images and the related work. Section 2 describes four new features. Section 3 describes the classification structure as well as the feature selection. Experiment results are included in Section 4 and finally Section 5 is the conclusion of the paper.
  \begin{figure}[!h]
     \centering 
     \setlength{\abovecaptionskip}{0.2cm}
      \includegraphics[width=0.99\columnwidth]{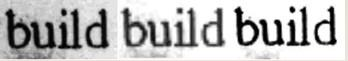}
    \caption{Image quality of original image/ picture mode image/ text mode image.}
  \end{figure}
  
\subsection{1.2 Five Types of Images}

We classify all images into five classes which are shown in Figure 2: Text, Picture, Mixed, Receipt and Highlight to reach optimal printing quality. And misclassification may cause image quality degradation. Text images contain only text. Picture images contain only pictures. Mixed images contain both text and pictures. Receipt images contain scanned receipts and Highlight images contain highlighted text or pictures. All the input images are scanned at 300 dots per inch (dpi) and the output result is one of the five classes.

Note that the receipt class and the highlight class are newly added. Receipt class represents a type of document images which have very low contrast and faded text which is shown in Figure 2(d). Its corresponding processing pipeline needs very strong contrast enhancement and text recovery to guarantee the readability of the copy\cite{bib3}\cite{bib4}. Compared with example given in Figure 1, receipt type needs even stronger enhancement for better quality. Highlight class typically contains colors that are more saturated compared with color in the natural images. An example is people using fluorescent pen to highlight text line in a document image. Highlighters usually produce very saturated color in order to attract the attention of readers. However, these saturated colors\cite{bib5}\cite{bib6} are sometimes difficult to be captured by the scanning devices. Or they are sensed as less saturated colors by the scanner. Figure 2(e) shows an captured image with highlight colors, which illustrates the vision of our device when sensing those colors. We can tell the colors in the upper half of the image are very weak from the view of scanners which is not desirable, but colors in the lower half are much more visible. So if we can tell the input image is a highlighted text document, we can produce more saturated color when generating output to match the appearance of actual highlighter ink. This can improve the user experience when copying highlighted document pages for better readability while at the same time preserving important area in the page.

   \begin{figure}[!h]
   \centering
    \subfloat[Sample text image.]{
      \includegraphics[width=0.3\columnwidth]{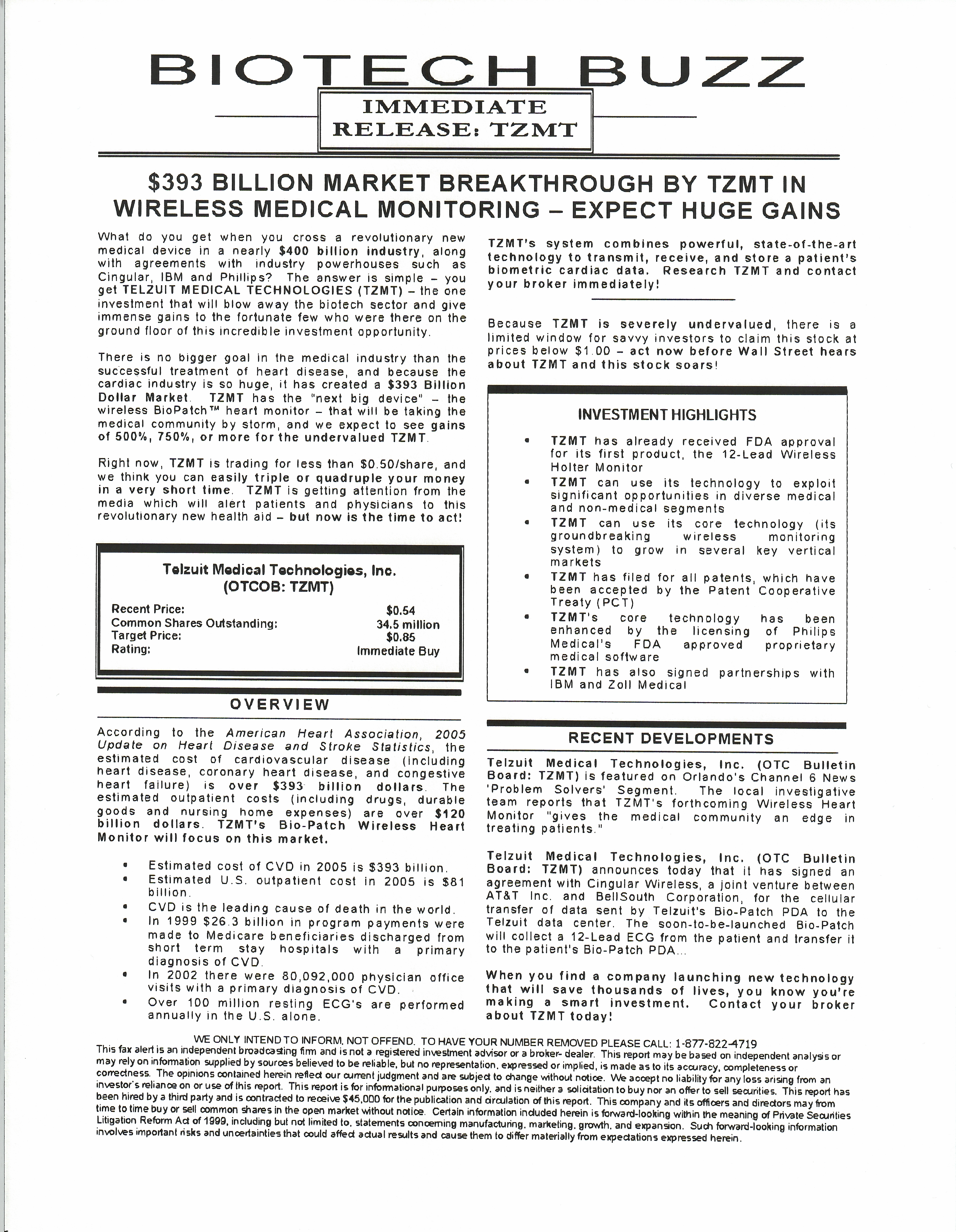}
    }
    \hfill
    \subfloat[Sample picture image.]{
      \includegraphics[width=0.3\columnwidth]{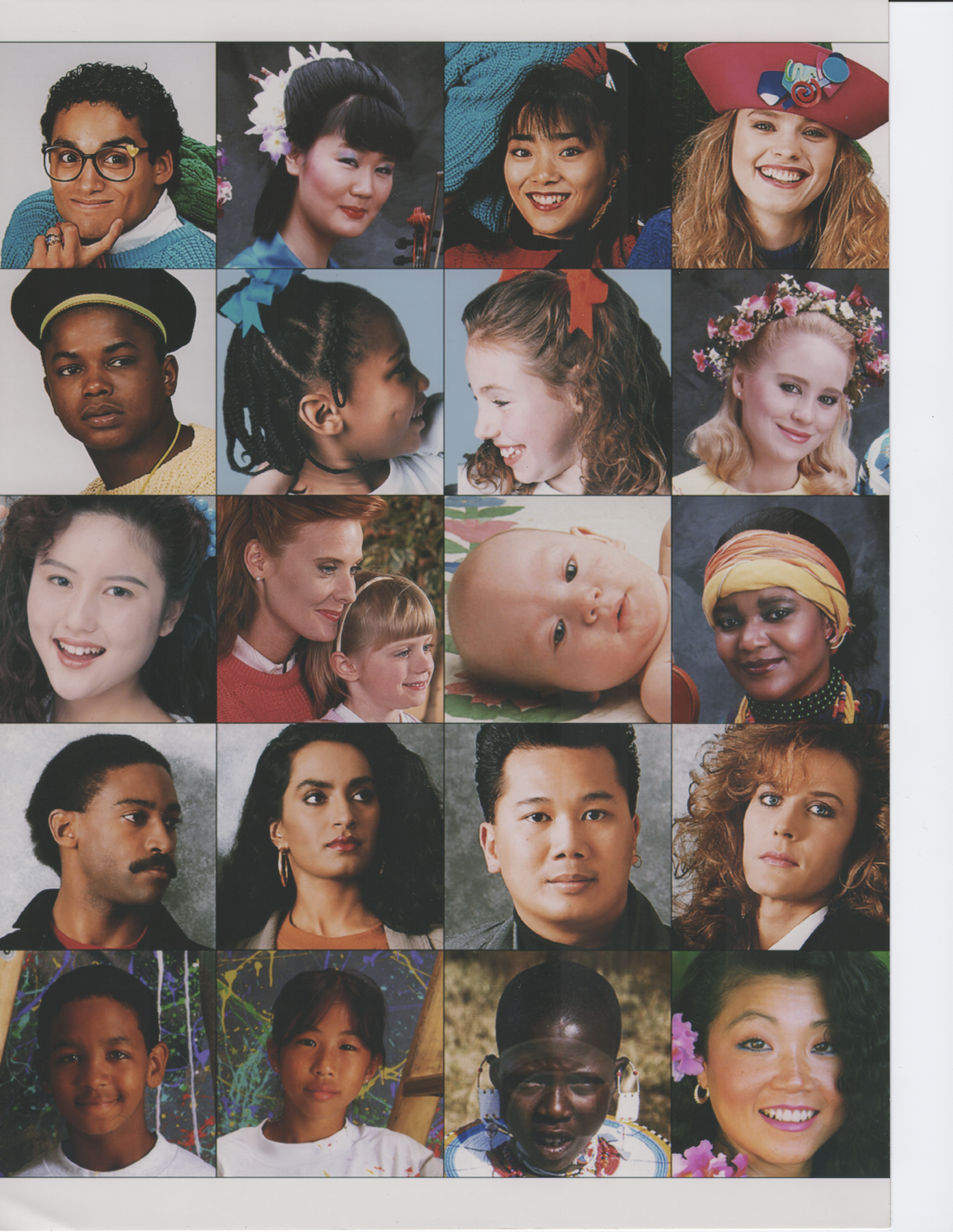}
    }
    \hfill
    \subfloat[Sample mixed image.]{
      \includegraphics[width=0.3\columnwidth]{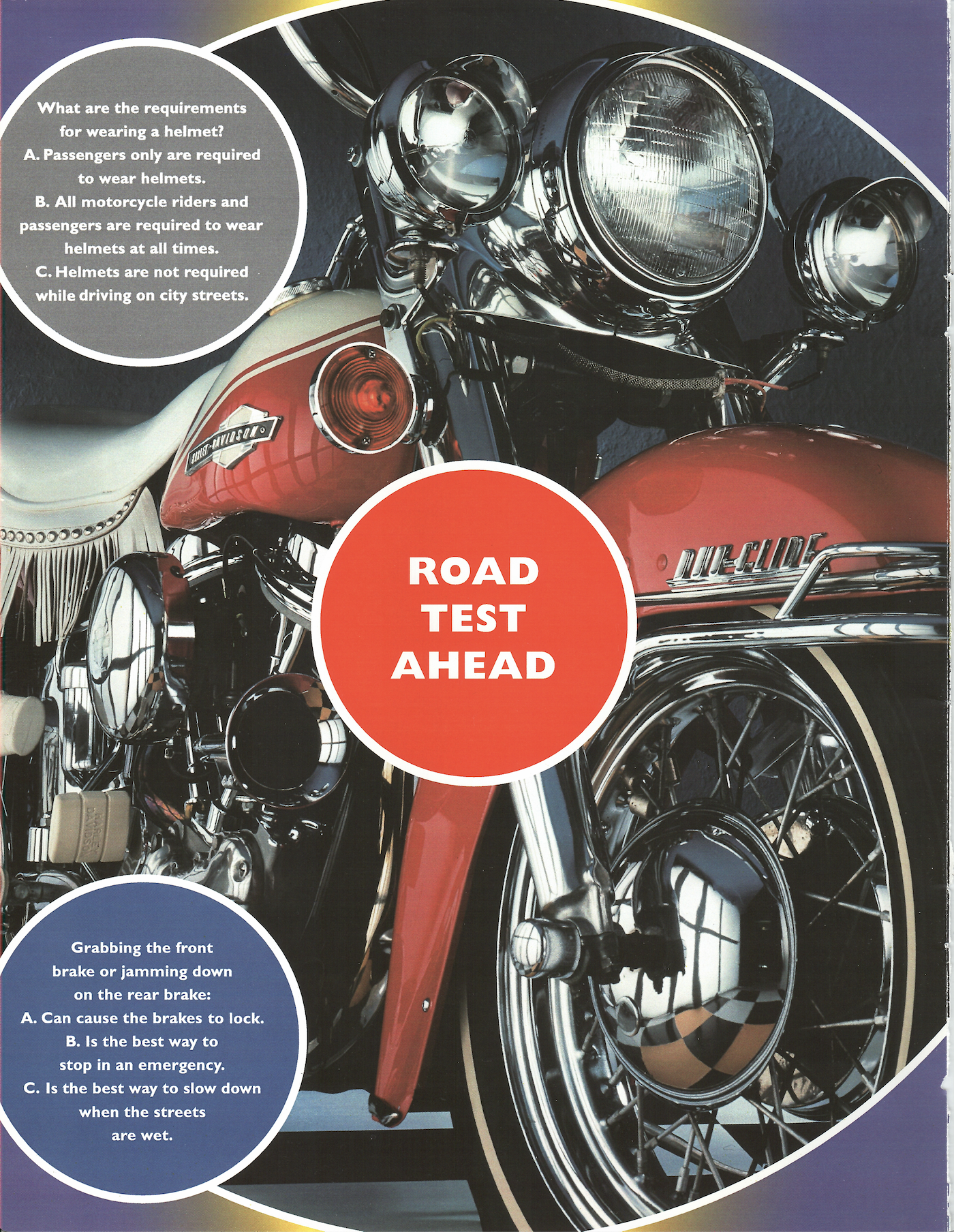}
    }
    \\
       \centering
    \subfloat[Sample receipt image.]{
      \includegraphics[width=0.3\columnwidth]{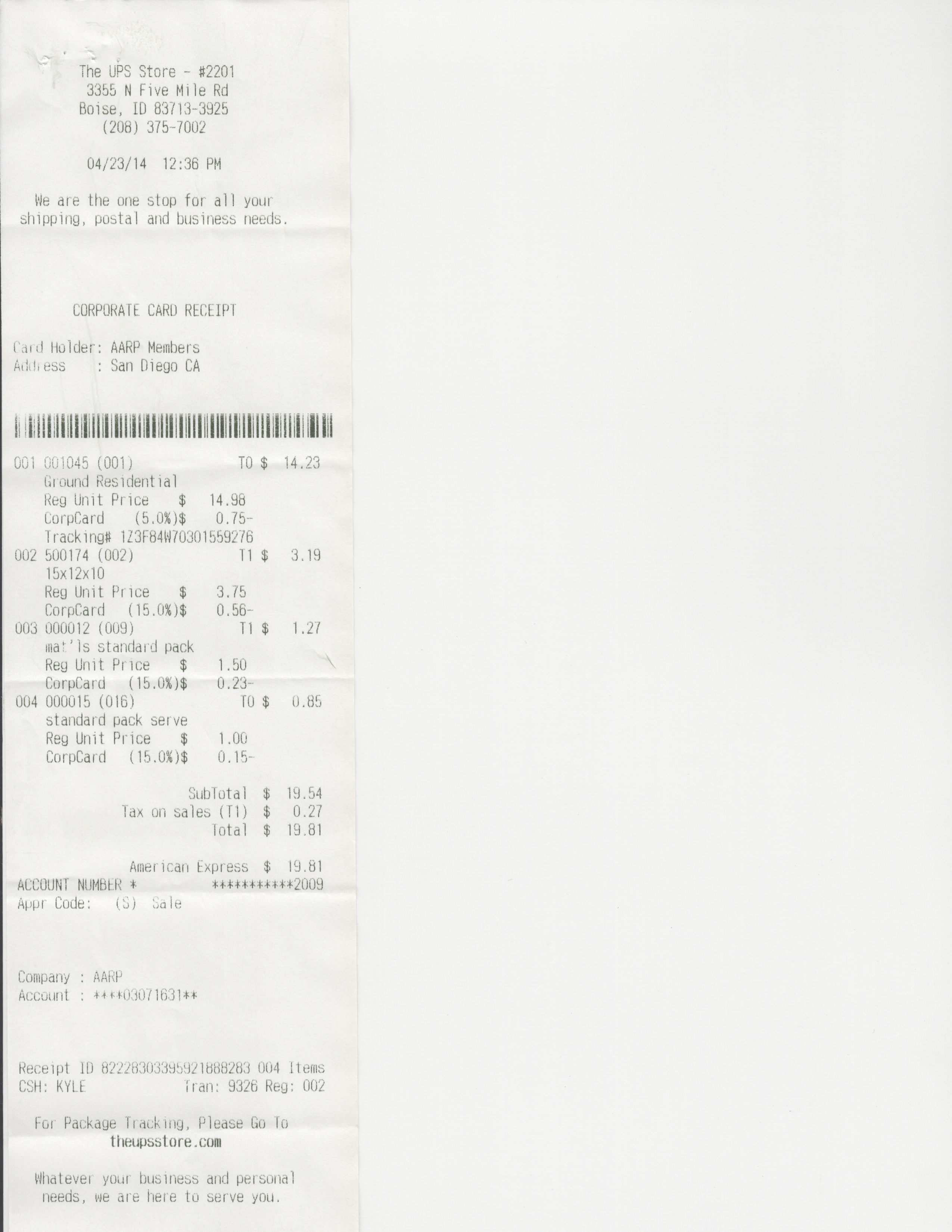}
    }
       \centering
    \subfloat[Sample highlight image.]{
      \includegraphics[width=0.3\columnwidth]{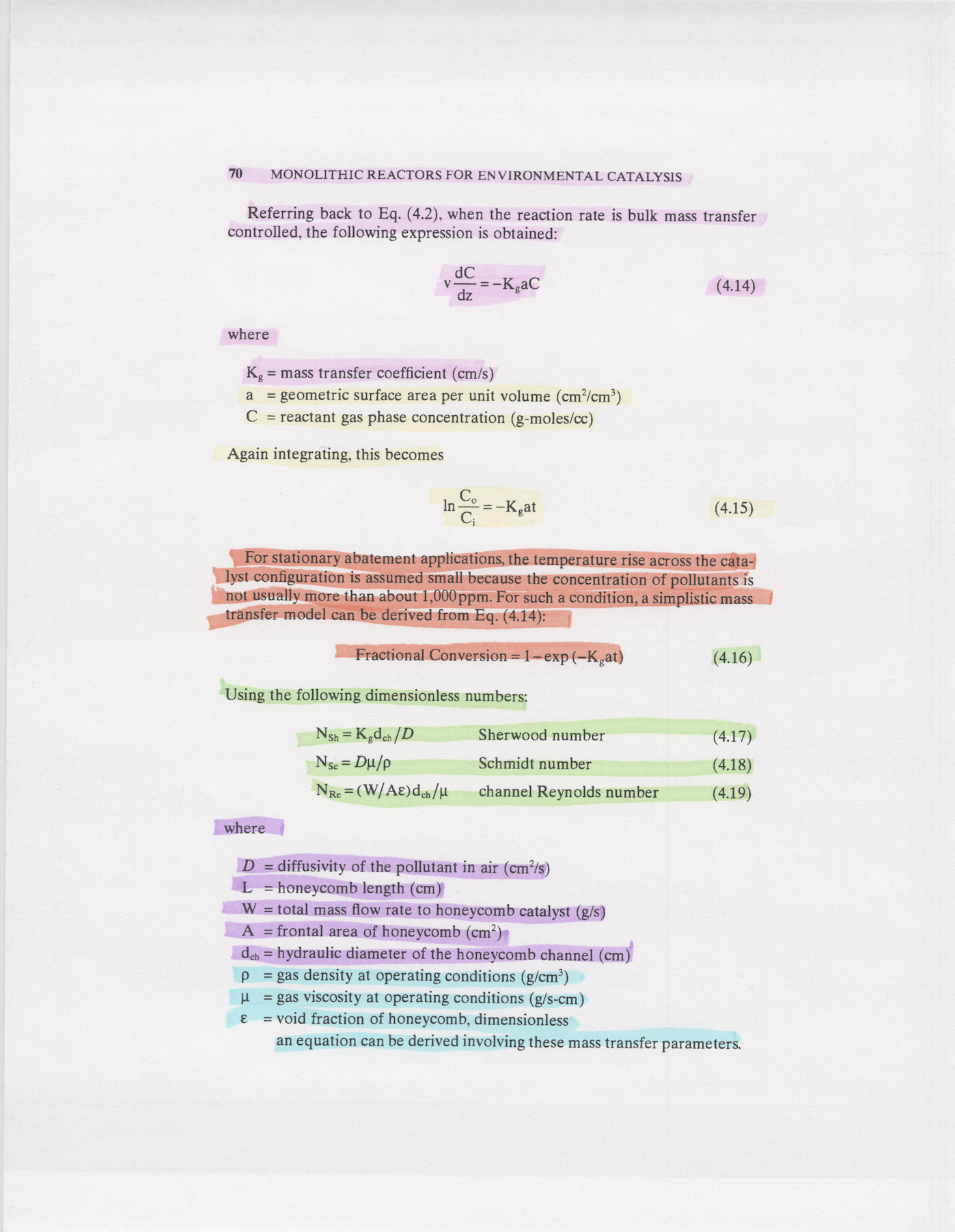}
    }
    \caption{Five types of images.}
  \end{figure}
  
\subsection{1.3 Related Work}

There has been a significant amount of research work done by Cheng Lu from our group which is related with an SVM-based classification\cite{bib7} of three types of images: text, picture and mixed. And he has already developed four features which are Histogram Flatness Score, Color Variability Score, Text Edge Count and Text Color Variance to train the classifiers. For the first feature which is the Histogram Flatness Score, typically, the histogram for a text image tends to have sharper and narrower peaks than the histogram for a non-text image such as a picture image. For the second feature which is the Color Variability Score, it is reasonable to assume that the non-text region of a text document contains only a few gray level values. So we build a block-mean histogram for the image to calculate a Color Variability Score. For the third feature which is the Text Edge Count, we calculate the number of text edges in an image based on the fact that the number of text edges in a text image is more than it is in a picture image. And for the last feature which is the Text Color Variance, we assume that text pixels in one image should have close luminance intensity values. Large variance in a certain area suggests that it’s a non-text region\cite{bib1}.

The new features are developed based on these four old features.

\section{2.0 Features}
\subsection{2.1 Chroma Histogram Flatness}

Although both natural images and highlighted text have chroma information, a very significant difference between natural color image and highlight text is that
highlighted text tends to include only one or few color while natural images have
richer color information. This can be illustrated in Figure 3 where we can find only
pink in highlighted text image while chroma in natural image is very rich. Then we
need a numerical value to represents this difference.

In Section 1.3, we introduce the Histogram Flatness Score in the luminance channel and it assumes that a text image has more peaky histogram. Similarly, if we build a histogram for a highlighted text in chroma space, we can expect that there is a single or few peaks, while the histogram for a natural image is more flat. Different from luminance histogram, chroma histogram is two-dimensional. Since our input image can be either RGB or LCH color space, we need to build histograms correspondingly. For RGB input, we transform it into YUV space and build a histogram on UV plane. For LCH input, we can build a histogram on CH space directly. As shown in Figure 4(a), UV space is in Cartesian coordinate system so we can uniformly partition it into 8$\times$8 areas and build a histogram based on that. However, CH space is described in polar coordinate system so we need to partition it differently. We uniformly divide the hue and chroma into 8 segments respectively, and thus giving us 64 areas which are not uniform in terms of space. LCH partition for building histogram is shown in Figure 4(b). For every input image, we cut it into 32$\times$32-pixel blocks and build a histogram for pixels in the block $i$ which is denoted as h$_i$. The Chroma Histogram Flatness for
the block $i$ is denoted as f$_i$ which is shown below:

\begin{equation}
 \qquad \qquad \qquad \quad f_i = \frac{\sqrt[N]{\prod_{n=0}^{N-1} h_i(n)}}{\frac{\sum_{n=0}^{N-1} h_i(n)}{N}}
\end{equation}

which is the geometric mean over the arithmetic where $N$ = 60 for YUV and $N$ = 56 for LCH in our case. It is worth noticing that we do not use all the bins of histogram to calculate $f_i$. That is because we should only focus on the chroma flatness and exclude those areas which are close to gray. Highlight-text will also have very low flatness if we consider those gray pixels which correspond to black text and white background. For YUV space, we ignore the central 4 bins and for LCH space we ignore the central 8 bins that are inside the smallest circle.

The Chroma Histogram Flatness F for the entire image is define as maximum fi
of all blocks in the image:

\begin{equation}
\qquad \qquad \qquad \qquad F = max(f_i)
\end{equation}

Because we do not expect seeing flat histogram of any block in the highlighted text image.

  \begin{figure}[!hb]
    \subfloat[Natural image.]{
      \includegraphics[width=0.47\columnwidth]{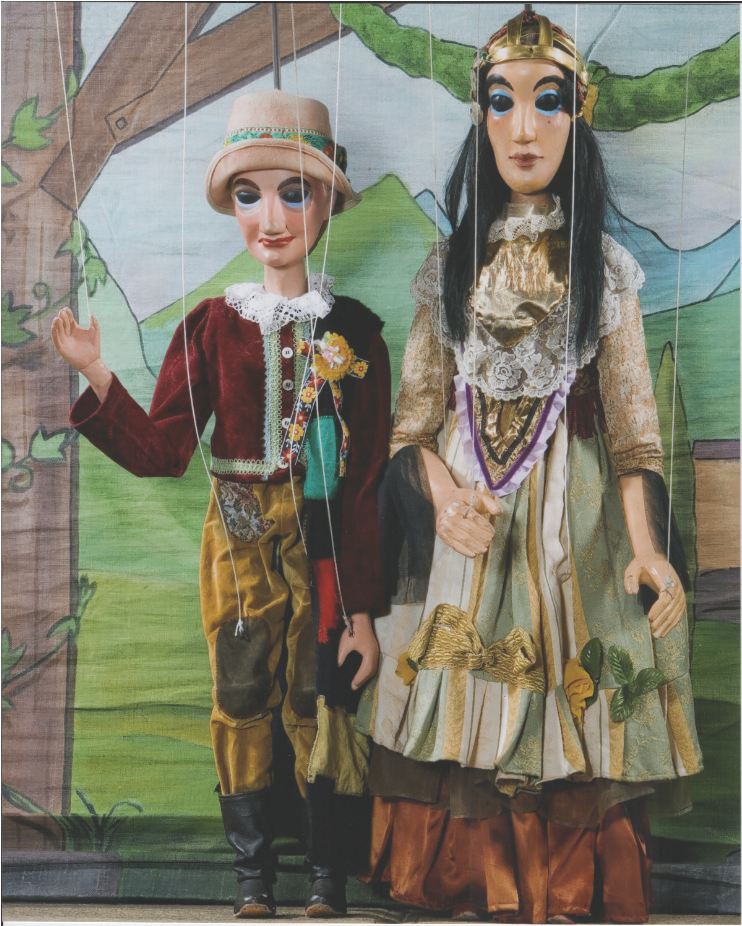}
    }
    \hfill
    \subfloat[Highlighted text.]{
      \includegraphics[width=0.452\columnwidth]{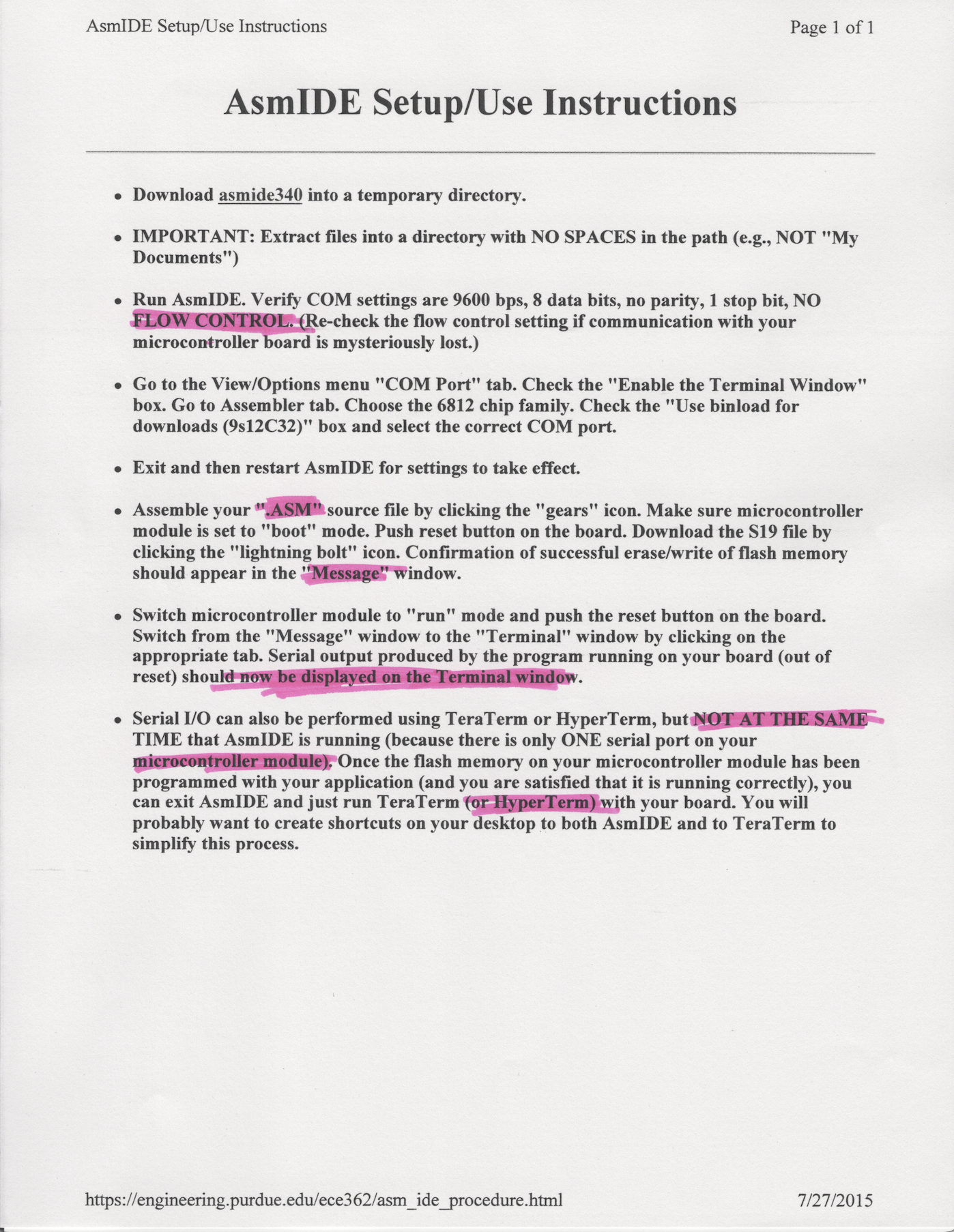}
    }
    \caption{Natural image v.s. Highlighted text.}
  \end{figure}
  
    \begin{figure}[!hb]
    \subfloat[YUV space partition.]{
      \includegraphics[width=0.47\columnwidth]{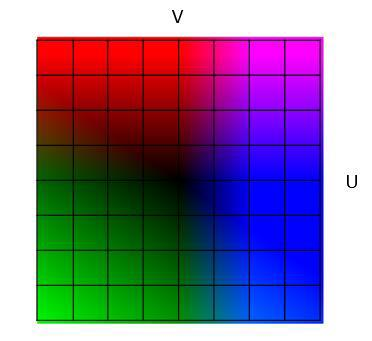}
    }
    \hfill
    \subfloat[LCH space partition.]{
      \includegraphics[width=0.5\columnwidth]{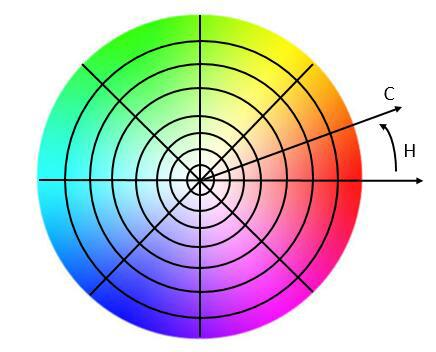}
    }
    \caption{LUV and LCH color space histogram.}
  \end{figure}

\subsection{2.2 Chroma Around Text}

On top of Chroma Histogram Flatness, we design another feature to detect chroma information of highlighted text image. Typically, we can expect that people use highlighter to emphasize text information, which means text strokes are usually covered and surrounded by highlight colors as shown in Figure 5. However for natural images, chroma information does not necessarily exists around edges. Given the fact, we can try to detect if there is chroma existence along the text edges in the image.

  \begin{figure}[!h]
     \centering 
     \setlength{\abovecaptionskip}{0.2cm}
      \includegraphics[width=0.99\columnwidth]{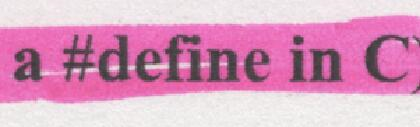}
    \caption{Highlighted colors around text.}
  \end{figure}

To find chroma around text strokes, we first need to find text edges in the image blocks. We follow the method introduced in Sec. 1.3 to find text edges. However, we need to note that reverse contrast text should not be considered any more. That is because it is rare that people mark light text with darker highlight which will cause poor readability. After finding a text edge, we search along its luminance increase direction to find if there is any chroma existence. Then we calculate chroma strength $(c)$ for two pixels along the luminance increase direction outside of text edge. In YUV space, chroma strength of a target pixel at position of (m, n) is defined as:

\begin{equation}
\qquad \qquad \qquad c(m, n) = u(m, n) + v(m, n)
\end{equation}

Note that we use simple summation of u and v here as approximation for faster
computation. In case of LCH space, chroma strength is equivalent to $c(m, n)$. The
process can be illustrated in Figure 6 after finding a text edge.
  
    \begin{figure}[!hb]
    \setlength{\belowcaptionskip}{0.3cm}
    \subfloat[Horizontal direction.]{
      \includegraphics[width=0.47\columnwidth]{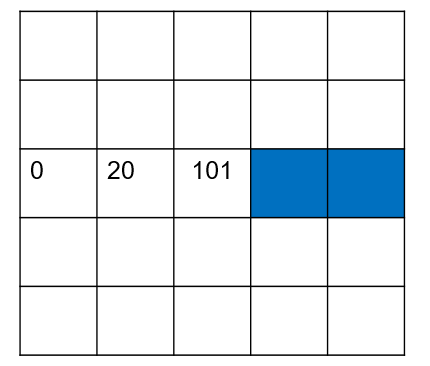}
    }
    \hfill
    \subfloat[Vertical direction.]{
      \includegraphics[width=0.47\columnwidth]{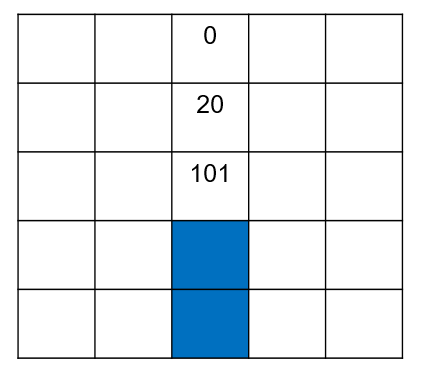}
    }
    \caption{Calculate c(m, n) of two pixels which are cover by blue.}
  \end{figure}

Similarly, we cut the input image into blocks with 32$\times$32 pixels and then find
$c(m, n)$ for all pixels outside text edges. Chroma Around Text $(c_i)$ of a block $i$ is
defined as:

\begin{equation}
\qquad \qquad \qquad \quad c_i = \frac{mean(c(m, n))}{std(c(m, n))}
\end{equation}

We also consider standard deviation of $c(m, n)$ because highlight should be consistent in a single block. Small $std(c(m, n))$ indicates that this block is more likely to contain highlight color. Similarly, chroma around text $(C)$ of an image is defined as the maximum value of $c_i$ of all blocks:

\begin{equation}
\qquad \qquad \qquad \qquad C = max(c_i)
\end{equation}

\subsection{2.3 Color Block Ratio}

Chroma Flatness and Chroma Around Text together can provide good discriminative power for detecting highlight image. However, they are not capable of handling text image with color background. One such example is given in Fig. 3.7. According to our feature design in Sec. 2.1 and Sec. 2.2, both features would strongly indicate that Figure 7 should be classified as highlighted text image. It is because both features only focus on local chroma and ignore global information.

  \begin{figure}[!h]
     \centering 
      \includegraphics[width=0.99\columnwidth]{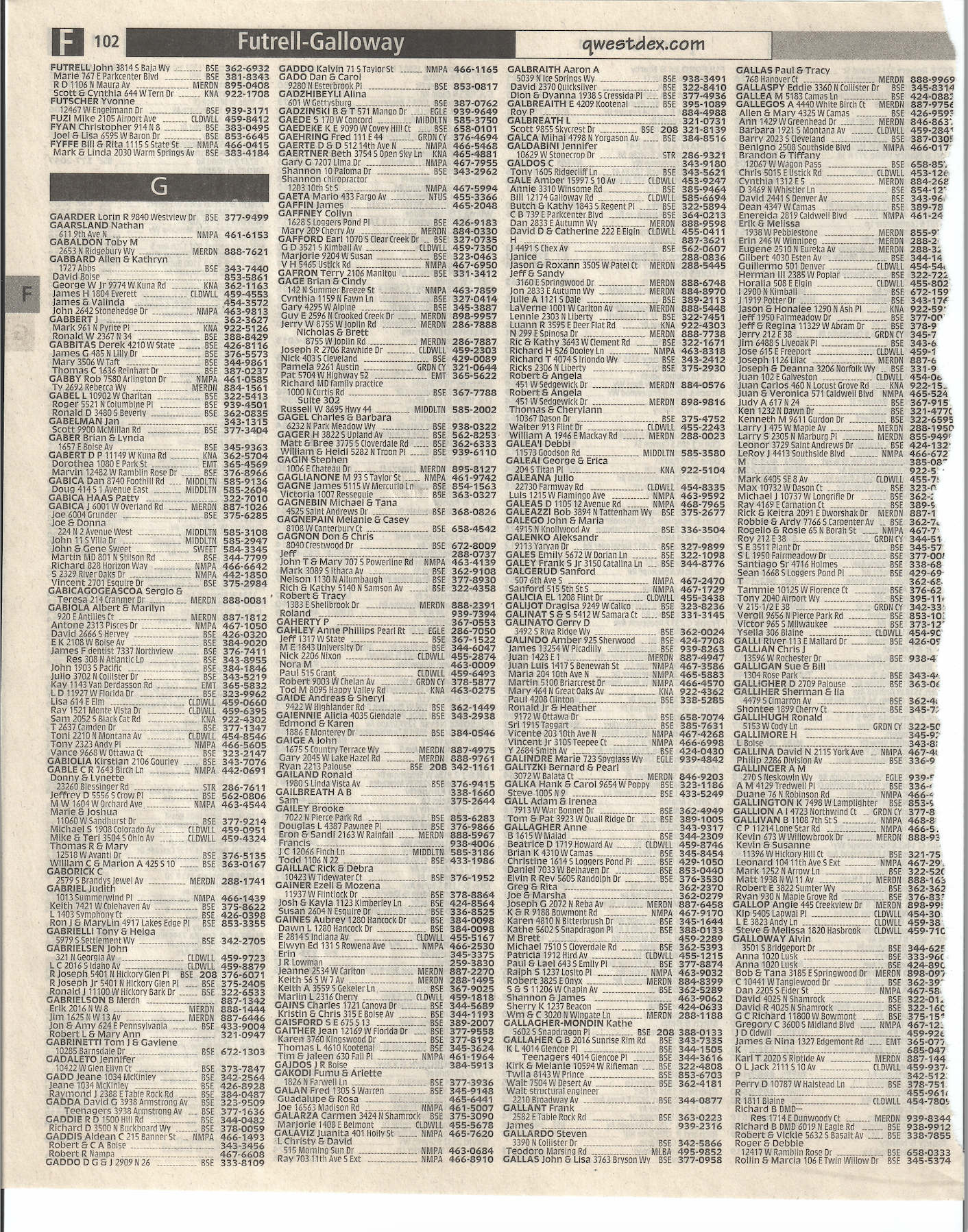}
    \caption{Text with yellow background.}
  \end{figure}

To address this problem, we introduce another feature called Color Block Ratio. We calculate the number of color pixels in every 32$\times$32-pixel block. A pixel is defined as color if its chroma strength is greater than a threshold value $T_c$ = 10. The chroma strength is defined in Equation 6 for YUV space. In LCH space is simply C channel. A block i is considered a color block if 10\% of its pixels are colored. We set $m_i$ = 1 if the block is colored otherwise $m_i$ = 0. Color block ratio $(R_c)$ of an image is defined as:

\begin{equation}
\qquad \qquad \qquad \qquad \quad R_c = \frac{\sum m_i}{\sum i}
\end{equation}

\subsection{2.4 White Block Ratio}

Compared with text image, receipt typically only occupies a small part of area on flat-bed scanner. An example is given in Figure 2 which shows that a scanned receipt only occupies upper-left corner of the plane. Utilizing this difference, we design the White Block Ratio feature. Again, we cut the input image into 32$\times$32-pixel blocks. For each block $i$, we check if 95\% of the pixels have luminance value which is larger than 230. If this is true, we consider this block to be a white block with $w_i$ = 1, otherwise we set $w_i$ = 0. White block ratio $(R_w)$ of an image is defined as:

\begin{equation}
\qquad \qquad \qquad \qquad \quad R_w = \frac{\sum w_i}{\sum i}
\end{equation}

\section{3.0 Classification}
\subsection{3.1 Classification Structure}

In order to make a more balanced multi-class classification and for easier tuning, we apply Directed Acyclic Graph-Support Vector Machine (DAG-SVM)\cite{bib8} to solve the problem. DAG-SVM is a tree-structured classification method that capable of making multi-class classification. Instead of making one vs rest decision, it tentatively make one vs one decision which allows more judgement from lower nodes. DAG-SVM classifiers that we use are shown in Figure 8:

  \begin{figure}[!h]
     \centering 
     \setlength{\abovecaptionskip}{0.2cm}
      \includegraphics[width=0.99\columnwidth]{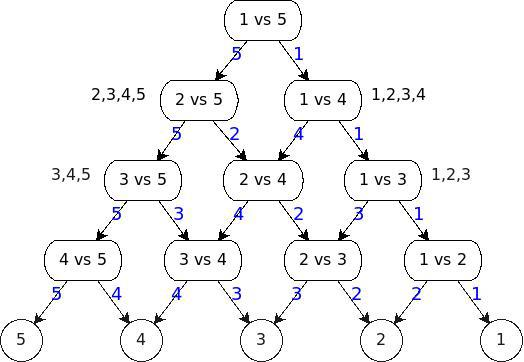}
    \caption{Tree structure of DAG-SVM.}
  \end{figure}
  
In our case, class 1 stands for mix, class 2 stands for text, class 3 stands for picture, class 4 stands for receipt and finally class 5 stands for highlight. It first decides if the input image is non-mix or non-highlight. If the image is classified as non-mix, then it is sent to the right node, otherwise it is sent to the left node. And the whole process goes from the top all the way to the bottom of the tree-structure. 
  
\subsection{3.2 Feature Selection}

We introduce four new features to classify receipt and highlight. Together
with the features we previously developed, we can represent each image $k$ with a
feature vector $f_k \in R^8$. However, we need to evaluate the contribution of each feature to the classification. To test the impact of each feature on the overall classification accuracy and time consumption, we adopt the leave-one-out feature selection. 

Because misclassifications are not equally weighted, we need to first consider the metric for feature selection. We define the weighted misclassification rate $(W_m)$ as below:

\begin{equation}
\qquad \qquad \qquad W_m = \frac{\sum_{i, j} w(i, j)n(i, j)}{\sum_j n(i, j)}
\end{equation}

where $w(i, j)$ is the weight of misclassification given in Table 2 and $n(i, j)$ is the number of images in this corresponding entry. To avoid biased measure due to different size of image set, the weighted sum needs to be normalized by the size of image set of each type. In our case, we have $\sum_j n(i, j) = 100$ for all the five types. But for completeness, we still keep the normalization term in Equation 8.

We first evaluate $M_m^8$ when we use all 8 features. Then we drop each feature $d$ at a time and evaluate $M_m^{\overline{d}}$. Then the feature impact factor for $d$ is defined as:

\begin{equation}
\qquad \qquad \qquad \qquad I_d = \frac{M_m^{\overline{d}} - M_m^8}{M_m^8}
\end{equation}

According to the Equation 9, the feature with larger $I_d$ is more important to our classification. Also, we measure the time consumption for each feature and take average over all the images. The results are shown in Table 1:

\begin{table*}[!h]
\centering
\renewcommand\arraystretch{1.5}
\label{my-label}
\caption{Feature impact factor and time consumption.}
\begin{tabular}{| p{0.18\columnwidth} | p{0.18\columnwidth} | p{0.18\columnwidth} | p{0.18\columnwidth} | p{0.18\columnwidth}| p{0.18\columnwidth} | p{0.18\columnwidth} | p{0.18\columnwidth} | p{0.18\columnwidth} |}
\hline
 \ Feature  & Histogram flatness & Color variability & Text edge count & Text color variance & Chroma around text & Chroma histogram flatness & White block ratio & Color block ratio   \\ \hline
 \ time \ (ms) & 12.01 & 12.51 & 14.97 & 73.92 & 36.12 & 11.45 & 0.67 & 0.81    \\ \hline
 \qquad $I_d$ & 24.61\% & 13.84\% & 11.02\% & 1.9\% & 10.2\% & 3.4\% & 48.43\% & 13.65\%    \\ \hline
\end{tabular}
\end{table*}

According to Table 1, we find Text Color Variability has the smallest impact
on overall classification accuracy and it consumes most of the time. Subsequently, we exclude this feature from our application and the final feature vector of an image $k$ is a 7-dimensional vector $f_k \in R^7$. This $f_k$ serves as the input to the DAG-SVM classifiers discussed in Section 3.1.

\section{4.0 Experimental Results}

To test the performance of our designed system, we build an image set which
includes 500 images scanned by flat-bed scanner. Each type of images has 100
images.

Different misclassifications are weighted differently due to its impact to image quality. Some misclassifications lead to larger image quality degradation and they should be assigned larger weight. Some other misclassifications cause smaller impact on image quality and should be assigned lower weight. The weight table of misclassification is shown in Table 2:

\begin{table}[!h]
\centering
\renewcommand\arraystretch{1.5}
\label{my-label}
\caption{Weights of misclassifications $w(i, j)$.}
\begin{tabular}{| p{0.14\columnwidth} | p{0.10\columnwidth} | p{0.10\columnwidth} | p{0.11\columnwidth} | p{0.12\columnwidth}| p{0.13\columnwidth} |}
\hline
 Ground truth  & Mix & Text & Picture & Receipt & Highlight   \\ \hline
 Mix & 0 & 3 & 5 & 6 & 4    \\ \hline
 Text & 3 & 0 & 10 & 6 & 2   \\ \hline
 Picture & 3 & 10 & 0 & 10 & 15    \\ \hline
 Receipt & 6 & 8 & 3 & 0 & 8   \\ \hline
 Highlight & 10 & 10 & 10 & 10 & 0    \\ \hline
\end{tabular}
\end{table}

To test the performance of our proposed algorithm, we conduct leave-one-out cross validation. We use rbf kernel with $\sigma$ for every node in Figure 8. Then we do exhaustive search for $\sigma$ and box constraint $C$ to find the best combination that produces the best cross validation result. The goodness of cross-validation result is measured by weight misclassification rate $(W_m)$ which is defined in Equation 8. When $W_m = W_m^*$ reaches minimum, the corresponding confusion matrix $n(i, j)$ in different color spaces are given in Table 3 and Table 4:
\\

\begin{table}[!h]
\centering
\renewcommand\arraystretch{1.5}
\label{my-label}
\caption{Confusion matrix $n(i, j)$ in YUV space.}
\begin{tabular}{| p{0.14\columnwidth} | p{0.10\columnwidth} | p{0.10\columnwidth} | p{0.11\columnwidth} | p{0.12\columnwidth}| p{0.13\columnwidth} |}
\hline
 Ground truth  & Mix & Text & Picture & Receipt & Highlight   \\ \hline
 Mix & 78 & 5 & 11 & 2 & 4    \\ \hline
 Text & 3 & 68 & 0 & 4 & 25   \\ \hline
 Picture & 5 & 0 & 94 & 1 & 0    \\ \hline
 Receipt & 0 & 3 & 3 & 88 & 6   \\ \hline
 Highlight & 4 & 8 & 1 & 5 & 82    \\ \hline
\end{tabular}
\end{table}

\begin{table}[!h]
\centering
\renewcommand\arraystretch{1.5}
\label{my-label}
\caption{Confusion matrix $n(i, j)$ in LCH space.}
\begin{tabular}{| p{0.14\columnwidth} | p{0.10\columnwidth} | p{0.10\columnwidth} | p{0.11\columnwidth} | p{0.12\columnwidth}| p{0.13\columnwidth} |}
\hline
 Ground truth  & Mix & Text & Picture & Receipt & Highlight   \\ \hline
 Mix & 76 & 6 & 12 & 1 & 5    \\ \hline
 Text & 4 & 72 & 0 & 9 & 15   \\ \hline
 Picture & 7 & 0 & 92 & 0 & 1    \\ \hline
 Receipt & 0 & 6 & 0 & 89 & 5   \\ \hline
 Highlight & 3 & 14 & 1 & 6 & 76    \\ \hline
\end{tabular}
\end{table}

\section{5.0 Conclusion}

In this paper, we introduce four new features to handle multi-class classification for AIO printer with scanning functionality. It extends the scope of the topic about the previous research work on SVM-based image classification of three types of images: text, picture and mix. Our proposed algorithm utilizes the chroma information of input image for better classification. Eventually, we use DAG-SVM with seven features to classify five classes of images which are text, picture, mixed, receipt and highlight. And based on the classification result, digital copiers or printers will produce images with better quality by choosing corresponding processing pipelines.

\end{document}